\documentclass[12pt]{elsarticle}

\usepackage{amssymb}
\usepackage{booktabs}
\usepackage{amsthm}
\usepackage{amsmath}
\usepackage{subcaption}
\usepackage{xcolor}
\usepackage{hyperref}
\def \Em{{\mathbb{E}}}
\def \Rm{{\mathbb{R}}}

\def \Ibf{{\mathbf I}}

\def \xbf{{\mathbf x}}

\def \zbf{{\mathbf z}}

\def \0bf{{\mathbf 0}}

\def \Rm{\mathbb{R}}

\def \Ncal{{\mathcal N}}

\def \Lcal{{\mathcal L}}

\usepackage{lineno}
\usepackage{enumitem}
\journal{Ecological Informatics}

\begin{document}

\begin{frontmatter}



\title{Dynamic $\beta$-VAEs for quantifying biodiversity by clustering optically recorded insect signals}

\author[inst1,inst2]{Klas Rydhmer}
\affiliation[inst1]{organization={Department of Geosciences \& Natural Resource Management, University of Copenhagen},%Department and Organization
            city={Copenhagen},
            country={Denmark}}
\affiliation[inst2]{organization={FaunaPhotonics A/S},%Department and Organization
            city={Copenhagen},
            country={Denmark}}

\author[inst3,inst4]{Raghavendra Selvan}
\affiliation[inst3]{organization={Department of Computer Science, University of Copenhagen},%Department and Organization
            city={Copenhagen},
            country={Denmark}}
\affiliation[inst4]{organization={Department of Neuroscience, University of Copenhagen},%Department and Organization
            city={Copenhagen},
            country={Denmark}}
\begin{abstract}
While insects are the largest and most diverse group of terrestrial animals, constituting ca. $80$\% of all known species, they are difficult to study due to their small size and similarity between species. Conventional monitoring techniques depend on time consuming trapping methods and tedious microscope-based work by skilled experts in order to identify the caught insect specimen at species, or even family level. Researchers and policy makers are in urgent need of a scalable monitoring tool in order to conserve biodiversity and secure human food production due to the rapid decline in insect numbers.

Novel automated optical monitoring equipment can record tens of thousands of insect observations in a single day and the ability to identify key targets at species level can be a vital tool for entomologists, biologists and agronomists. Recent work has aimed for a broader analysis using unsupervised clustering as a proxy for conventional biodiversity measures, such as species richness and species evenness, without actually identifying the species of the detected target.

In order to improve upon existing insect clustering methods, we propose an adaptive variant of the variational autoencoder (VAE) which is capable of clustering data by phylogenetic groups. The proposed {\em dynamic} $\beta$-VAE dynamically adapts the scaling of the reconstruction and regularization loss terms ($\beta$ value) yielding useful latent representations of the input data. We demonstrate the usefulness of the dynamic $\beta$-VAE on optically recorded insect signals from regions of southern Scandinavia to cluster unlabelled targets into possible species. We also demonstrate improved clustering performance in a semi-supervised setting using a small subset of labelled data. These experimental results, in both unsupervised- and semi-supervised settings, with the dynamic $\beta$-VAE are promising and, in the near future, can be deployed to monitor insects and conserve the rapidly declining insect biodiversity.\footnote{Source code available at: \url{https://github.com/remhdyr/dynamicBeta}}
\end{abstract}

% \includegraphics[width=0.99\textwidth]{figures/Graphical abstract.pdf}
% % \caption{Latent representation of optically recorded insect wingbeat frequency spectra. The proposed dynamic $\beta$-VAE is able to cluster unlabelled field recordings into compact clusters that correspond with the species labels, seen here as different colours for each of the 12 named species groups.}
%   \label{fig:graphical_abstract}

% \begin{highlights}
% \item First application of a Variational Auto Encoder (VAE) for biodiversity assessment of insect signals 
% \item The proposed dynamic $\beta$-VAE is capable of clustering optically recorded insect signals using a compact 2-d space
% \item The $\beta$-coefficient is dynamically adjusted during training to balance the trade-off between reconstruction and regularization terms
% \item A fully unsupervised model is outperforming conventional methods, such as PCA whereas a semi-supervised method improves upon the unsupervised model results even further
% \end{highlights}

\begin{keyword}
{unsupervised clustering, VAE, insect classification, biodiversity}
% %% PACS codes here, in the form: \PACS code \sep code
% %% MSC codes here, in the form: \MSC code \sep code
% %% or \MSC[2008] code \sep code (2000 is the default)
\end{keyword}

\end{frontmatter}

\section{Introduction}
Insects make up the majority of all known animal species with ca. 1 million described species and an estimated 3-4 million yet to be discovered ~\cite{may1988many, stork2018many}. While insects are numerous and found in almost all habitats, the total insect population is thought to be shrinking at an alarming rate. An influential report recently reported a $70$\% loss of flying insect biomass in 30 years \cite{hallmann2017more}. These losses have mainly been driven by changes in the agricultural landscape, increased use of pesticides and the spread of disease, but the exact reasons and consequences are still unknown \cite{potts2016assessment, goulson2015bee}. In order to accurately measure the biodiversity and health of the insect community across various biotopes (or habitats), researchers, agronomists, policy makers and institutions are in need of insect monitoring capabilities from multiple areas, over long periods of time.

Conventional insect biodiversity monitoring typically involves various trapping methods, each with their own bias towards different species, which makes it difficult to compare the results across studies \cite{muirhead2012trap}. The collected insect specimens are further identified under microscopes by highly trained experts. These methods provide data with very high specificity but are time consuming and expensive which severely limits the ability to collect data on a large scale, or over extended time periods, with high spatial and temporal resolution.

In recent years, new technologies have been developed for insect monitoring such as automated traps \cite{potamitis2014electronic, potamitis2017automated}, acoustic methods \cite{4218202, mankin2011perspective} and optical instruments such as the entomological lidar \cite{brydegaard2014advantages, jansson2018passive, shaw2005polarization, genoud2017remote}. In general, these methods provide large amounts of data with a high temporal resolution but with lower specificity {compared to conventional methods} \cite{kirkeby11advances, fanioudakis2018mosquito, potamitis2017automated, chen2014flying}. The introduction of automated and continuous monitoring methods has the potential to greatly improve biodiversity monitoring and, consequently, conservation efforts. In order to utilize the full potential of these new methods, large number of unlabelled insect recordings need to be translated into a quantifiable biodiversity index, comparable to conventional estimates.

% Overview of paper goes here?
In this work, we combine the rich data from lidar entomology with the powerful capabilities of variational auto-encoders (VAEs) \cite{kingma2014auto}. A common trade-off that is difficult to achieve in VAEs is between its two loss components: reconstruction loss and regularization loss. Balancing these two components can yield useful low-dimensional features (representations) of the input data which can be further analyzed to perform clustering of the input data \cite{bengio2013representation,higgins2016beta}. 
We introduce a dynamically changing formulation of the scaling of the loss terms ($\beta$). The proposed $\beta$ dynamics takes instantaneous changes to the loss terms and the  historical performance during training into account to keep both the reconstruction and regularization performance near their optimum.

The proposed {\em dynamic} $\beta$-VAE is trained on unlabelled insect recordings collected using a novel optical insect sensor at several sites in southern Scandinavia. We demonstrate its ability to successfully extract features from input data into the  regularized latent space, and to cluster the data into an appropriate number of clusters. We experimentally validate improvements compared to more conventional methods such as the hierarchical clustering algorithm (HCA) \cite{brydegaard2020lidar, kouakou2020entomological} and principal component analysis (PCA). Additionally, we show that the introduction of a semi-supervised data set further improves the clustering performance on the unlabelled data when evaluated on known phylogenetic groups.

\section{Background \& Related Work}
\subsection{Lidar-entomology \& clustering}
% Less about features here and more about clusters! <- Actually!
Lidar entomology is an insect monitoring method where insects are recorded as they enter an infrared laser beam which can sometimes extend for kilometers. It might be the fastest way to record large amounts of insect data, yielding up to several tens of thousands of optically recorded insect signals per day \cite{brydegaard2020lidar}. This data consists of time series, where the signal intensity varies with the insect cross section and wing beats. As the wing-beat frequency (WBF) varies between insects groups, it can to some degree be used to distinguish between species, alone or with other extracted features \cite{kirkeby11advances, fanioudakis2018mosquito,
potamitis2017automated, chen2014flying, gebru2018multiband}.

While the ability to identify a number of key species from automated sensors would be greatly beneficial to the entomological community, it is not sufficient to quantify the biodiversity. Instead, the total number of species (species richness) and their relative distributions (species evenness) are the commonly used measurements. Previous efforts to derive these numbers from a large number of optically recorded insect signals have been made using HCA on the WBF power spectra \cite{brydegaard2020lidar, kouakou2020entomological}. However, in order to cover a broad range of frequencies with sufficient resolution, a high dimensional feature space is required and the distance measures are non-trivial. For reduced model complexity and improved computational and clustering performance, a reduction of the parameter space is desired.

% Auto encoders and entomology - Related work
In order to reduce the parameter space while retaining the necessary information, various algorithms for extracting the WBF and other physical properties from insect recordings have been proposed and used  \cite{kirkeby11advances, gebru2018multiband, qi2015effective, jansson2018first, li2020bark}. Machine learning based methods for feature extraction, such as auto-encoders (AE), have also been used to extract additional features \cite{qi2015effective}, and very recently, to cluster acoustically recorded bird songs \cite{rowe2021acoustic}. While an AE is able to generate high quality features for classification, a known behaviour of AE is the irregularity of the latent feature space where two similar data inputs might end up with very different latent representations. This makes the extracted features from an auto-encoder unsuitable for clustering recordings of similar insect species and quantifying the diversity of the recorded insects.

\subsection{VAEs and $\beta$-Annealing}
Variational autoencoders (VAEs) consist of a regularized probabilistic encoder-decoder pair and are some of the most powerful representation learning methods \cite{bengio2013representation,kingma2014auto}. They have seen broad applications in generative modelling and unsupervised learning tasks.

Given unlabelled input data consisting of $N$ samples with $F$ features, $\xbf \in \Rm^{N\times F}$, the probabilistic encoder of a VAE maps the input to the posterior density $p(\zbf|\xbf)$ over the latent variable, $\zbf \in \Rm^{N\times L}$. In practice, $L<<N$ and the encoder neural network approximates the true posterior density, $p(\zbf|\xbf)$, with a multivariate Gaussian, $q_\theta(\zbf|\xbf)\sim \Ncal(\mu_\theta,\sigma^2_\theta)$. The decoder of a VAE reconstructs the input data from the latent variable and is given by the density function $p_\phi(\xbf|\zbf)$. The encoder and decoder neural networks are parameterised by $\theta$ and $\phi$, respectively. The optimization objective of a VAE consists of two competing terms and it can be shown to be~\cite{kingma2014auto}
\begin{align}
    \mathcal{L}_{\text{VAE}} &=  -\Em_{q_{\theta}} \big [ \log p_\phi(\xbf|\zbf)\big] + \text{KL} \big[q_{\theta}(\zbf|\xbf) || p(\zbf)\big] \label{eq:vae}\\
    \mathcal{L}_{\text{VAE}} &\triangleq \mathcal{L}_{\text{rec}} +  \mathcal{L}_{\text{reg}} \label{eq:vae_objective}
\end{align}
The quality of the auto-encoded reconstructions is controlled by the reconstruction loss $\mathcal{L}_{\text{rec}}$, which is the first term in Eq.~\eqref{eq:vae}.  The encoder density is regularized to match the prior over the latent variable, $p(\zbf)\sim \Ncal(\0bf,\Ibf)$, enforced by the regularization loss, $\mathcal{L}_{\text{reg}}$, which is the Kullback-Leibler divergence (KLD) term in Eq.~\eqref{eq:vae}. At a high level, the regularization term controls the smoothness or the regularity of the latent space. Well structured and smooth latent spaces can yield useful representations of the input data. 

The trade-off between the two loss terms can have influence on the performance of any VAE. A VAE where the reconstruction term dominates might be able to reconstruct the input data well with a latent space that might not be interesting for the downstream tasks (such as clustering). To alleviate this, a simple trick of scaling the regularization term $\Lcal_\text{rec}$
 was used in ~\cite{higgins2016beta} resulting in a modified objective:
\begin{equation}
    \mathcal{L}_{\beta\text{-VAE}} = \mathcal{L}_{\text{rec}} + \beta  \mathcal{L}_{\text{reg}}.    
    \label{eq:beta-vae}
\end{equation}
Here the role of $\beta \geq 0$ is to balance the reconstruction- and regularization losses. Typically, lower $\beta$ values yield better reconstructions but a less regularized latent space and less disentangled features. On the other hand, higher $\beta$ may lead to posterior collapse, where all reconstructions are reduced to the average input and the KLD approaches zero.
Various methods have been proposed to overcome this instability in achieving a reasonable trade-off between the loss terms. A common implementation is $\beta$-annealing, where $\beta$ is gradually increased from a very low value up to a fixed point. While this solves the initial stability problems, the task of finding the optimal value of $\beta$ remains. Recently, it has been shown that repeating the process with a cyclic $\beta$ can lead to better performance \cite{fu2019cyclical}. However, when unchecked both implementations face the risk of posterior collapse (vanishing KLD) once $\beta$ enters a stationary phase.

More recently, several approaches have attempted to {\em adapt} $\beta$ instead of using a fixed or scheduled scaling~\cite{shao2020controlvae,shao2020challenging,asperti2020balancing}. In the controlVAE formulation in \cite{shao2020controlvae}, rather than gradually increasing $\beta$ to a maximum value (annealing), it is controlled with feedback from a non-linear proportional-integral (PI) controller to keep the KLD at a desired level. This addresses the vanishing KLD problem but the users still have to set the desired value of the KLD, which might not be straightforward for many applications.

\section{Methods}
The primary goal of this work is to obtain low-dimensional latent representations suitable for clustering of the high-dimensional input data. Specifically, the objective is to obtain latent space encodings of the insect spectral data such that similar species of insects are positioned close to each other. To this end, we propose to dynamically adapt, the otherwise constant scaling factor, $\beta$ of a standard $\beta$-VAE.

In our proposed dynamic $\beta$-VAE formulation, the changes in reconstruction and regularisation losses are monitored throughout the training process; these changes are used to adapt the $\beta$ value in each epoch using a simple control algorithm. If either the reconstruction- or regularization losses increase above a specific level of their historical minimum then $\beta$ is adjusted (either by increasing or decreasing) until a new optimum is attained. This dynamic control of $\beta$ maintains a steady trade-off between the two loss terms while reducing the global loss function by latching on to the historical minimum of the loss components. 

In the remainder of this section, we detail the dynamic $\beta$-VAE and formulate a semi-supervised variant of the model using a new clustering loss component.

\begin{figure}[t]
    \includegraphics[width=0.9\columnwidth]{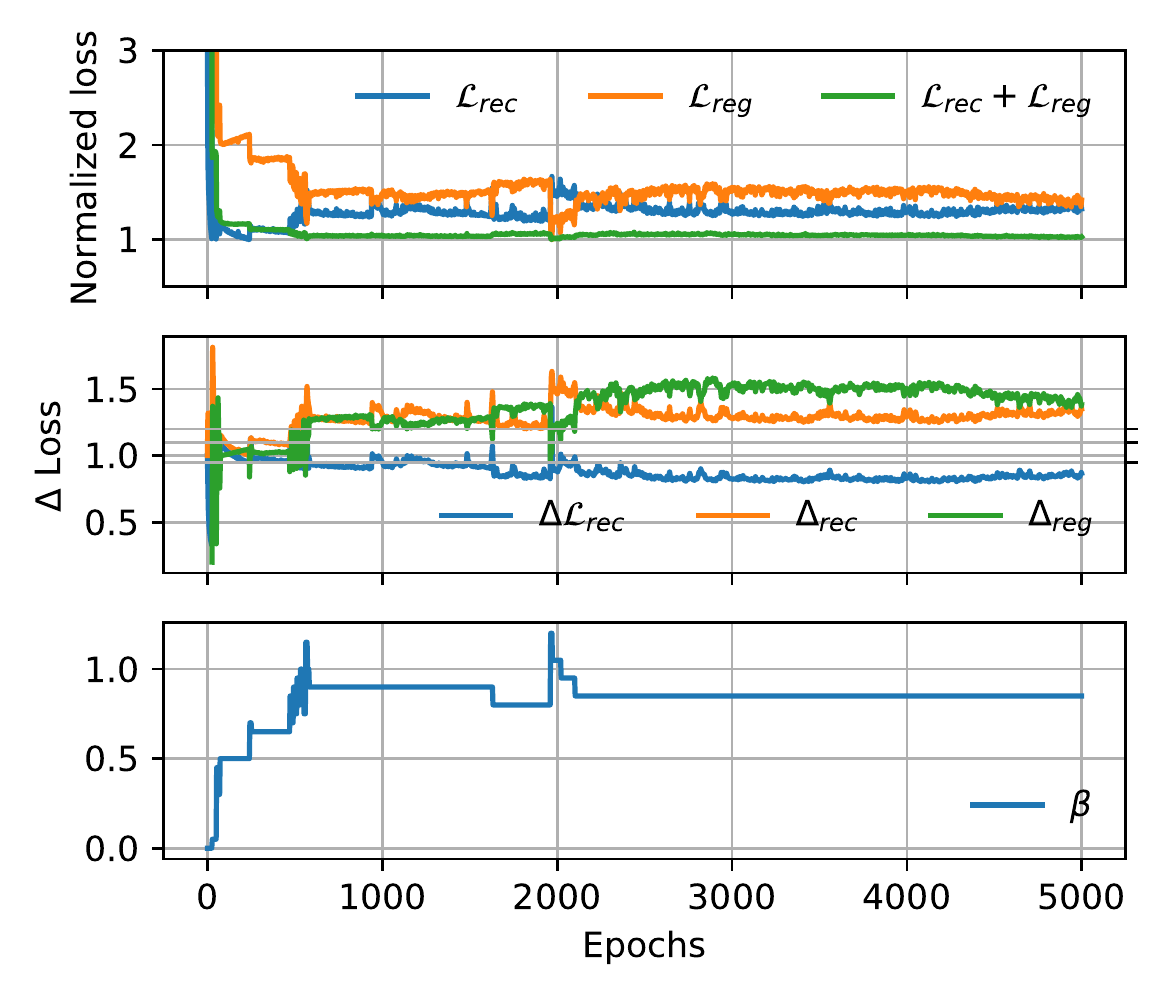}
    \vspace{-0.5cm}
    \caption{The evolution of $\beta$ during training of an unsupervised dynamic $\beta$ VAE. After a 25 epoch warm-up phase when $\beta=0$, it is dynamically adjusted based on $\Delta \mathcal{L}_{rec}$, $\Delta _{rec}$ and $\Delta_{reg}$. Thereby, $\mathcal{L}_{rec}$ and $\mathcal{L}_{reg}$ remain balanced without increasing the total loss (Row 1) implying stable model convergence.}
    \label{fig:progression}
    \vspace{-0.5cm}
\end{figure}

\subsection{Dynamic $\beta$-VAE}
\label{sec:dyn}
The key contribution in this work is an {online}, adaptive formulation of the $\beta$-VAE using dynamic control of $\beta$. This is achieved by varying $\beta$ at each epoch, based on the instantaneous changes in the reconstruction- and regularization terms in Eq.~\eqref{eq:beta-vae}, with an objective of not letting either of the loss terms to dominate the overall model optimization. This results in a trade-off between sufficiently good reconstructions and adequately regularized latent space yielding representations of the input data that are useful for the downstream task.

At any given epoch $t$ the objective for the dynamic $\beta$-VAE is given by,
\begin{equation}
    \mathcal{L}^{(t)} = \mathcal{L}_{\text{rec}}^{(t)} + \beta^{(t)}  \mathcal{L}_{\text{reg}}^{(t)}.
    \label{eq:dyn-vae}
    \end{equation}
The dynamically controlled $\beta^{(t)}$ is formulated using the signum function\footnote{The signum function, $\psi[x]$ returns the sign of any real number $s\in \Rm$
\begin{equation*}
    \psi[s] = \begin{cases}
    +1 & \quad s > 0 \\
     0 & \quad s = 0\\
    -1 & \quad s < 0
    \end{cases}
\end{equation*}}, $\psi[\cdot]$, given by
\begin{align}
    \beta^{(t)} = \beta^{(t-1)}  &- \frac{b}{4}\left(1- \psi\left[\Delta_\text{reg}\right] \right)  \left( 1+\psi\left[\Delta_\text{rec}\right]   + {\Delta\Lcal_\text{rec}} \right) \nonumber \\
    &+ \frac{a}{4} \left(1- \psi\left[\Delta_\text{rec}\right] \right)  \left( 1+\psi\left[\Delta_\text{reg}\right] - \Delta\Lcal_\text{rec} \right)
    \label{eq:dynamics}
\end{align}
where 
\begin{align}
    \Delta_\text{rec} &= \Lcal_\text{rec}^{(t)} - w_1 \min \left( \Lcal_\text{rec}^{(:t-1)}\right)  \label{eq:w1}\\
    \Delta_\text{reg} &=\Lcal_\text{reg}^{(t)} - w_2 \min \left( \Lcal_\text{reg}^{(:t-1)}\right)     \label{eq:w2}\\
    \Delta\Lcal_\text{rec} &=\psi\left[\Lcal_\text{rec}^{(t)} - w_3  \Lcal_\text{rec}^{(t^\prime)} \right] + \psi\left[ \Lcal_\text{rec}^{(t)} - w_4  \Lcal_\text{rec}^{(t^\prime)} \right]
    \label{eq:w3}
\end{align}

with hyperparameters $[a,b,w_1,w_2,w_3,w_4] \in \Rm^+$. The notation $(:t-1)$ is used to indicate all epochs up to (t-1) and $(t^\prime)$ is the epoch when $\beta$ was last changed. {The terms associated with $(:t-1)$ provide a form of long term memory of the previous local optima for each of the two loss terms}.

The $\beta$ dynamics in Eq.~\eqref{eq:dynamics} can be divided into two regimes aimed at optimizing reconstruction- and regularization terms corresponding to increase- and decrease of $\beta$, respectively. 
\\
{\bf Reconstruction regime} ($\beta \downarrow$):
The value of $\beta$ is decreased due to the $-b$ term in Eq.~\eqref{eq:dynamics} when  $\Delta_\text{rec}$ is positive; meaning the reconstruction loss is increasing compared to the historical minimum reconstruction loss, according to Eq.~\eqref{eq:w1}. The $\beta$ decrease rule also checks if the regularization loss is decreasing compared to the historical minimum with the term $1-\psi\left[\Delta_\text{rec}\right]$ in Eq.~\eqref{eq:dynamics}. 
\\
{\bf Regularization regime} ($\beta \uparrow$):
The increase in $\beta$ happens due to the $+a$ term in Equation~\eqref{eq:dynamics} when $\Delta_\text{reg}$ is positive; meaning the regularization loss is increasing according to Eq.~\eqref{eq:w2}. The increase rule checks if the reconstruction loss has decreased compared to the historical minimum with the term $1-\psi\left[\Delta_\text{reg}\right]$ in Eq.~\eqref{eq:dynamics}.

Additionally, $\Delta\Lcal_\text{rec}$ in Eq~\eqref{eq:w3} nudges a change in $\beta$ based on the last update to $\beta$. This allows $\beta$ to get out of plateaus of either stable reconstruction or regularization regimes. 

%% Figure / diagram of beta balancing goes here?
In Figure~\ref{fig:progression}, one instance of optimizing the dynamic $\beta$-VAE with the objective in Eq.~\eqref{eq:dyn-vae} is shown. The value of $\beta$ increases until about epoch 700 at which it plateaus and decreases (Figure~\ref{fig:progression}, row 3). At epoch 2000, it has finally stabilized. These changes are correlated with changes to $\mathcal{L}_{reg}$ and $\mathcal{L}_{rec}$ captured in the second row of Figure \ref{fig:progression}, estimated according to Equations \ref{eq:dynamics} to \ref{eq:w3}.

\subsection{Semi-supervised clustering}
Using a small subset of labelled data that optimizes a relevant loss could steer learning of representations that are more expressive for the downstream tasks under consideration. One approach to achieve this is to introduce auxiliary tasks based on the labelled data, resulting in a semi-supervised learning setup~\cite{figueroa2017learning}. 

As we are interested in clustering of insect species based on their latent representation, we enforce clustering of a small subset of labelled examples to improve the overall clustering. An additional loss term, $\Lcal_\text{cls}$, based on the auxiliary task is introduced to the model optimization:
\begin{equation}
    \mathcal{L} = \mathcal{L}_{\text{rec}} + \beta^{(t)} \mathcal{L}_{\text{reg}} + \gamma^{(t)}  \mathcal{L}_{\text{cls}}
    \label{eq:final_obj}
\end{equation}
where $\gamma^{(t)}$ is the scaling of the clustering loss component.

The clustering loss, $\mathcal{L}_{\text{cls}}$, has two components to encourage intra-class cohesion and inter-class repulsion. Intra-class cohesion is captured as the sum of distance between all data points belonging to a particular class and the corresponding cluster centroid location. Assuming $K$ clusters with centroids $\hat\zbf_k \in \Rm^L, k=1\dots K$ and $N_k$ cluster members with class label $k$ denoted $\zbf^k$, the intra-class cohesion distance is given by:
\begin{equation}
    d_C = \sum_{k=1}^K \sum_{i=1}^{N_k} d(\zbf_i^{k},\hat\zbf_k).
\end{equation}
The inter-class repulsion distance is captured using the sum of all pairwise distances between the cluster centroids:
\begin{equation}
    d_R = \sum_{i=1}^K\sum_{j<i} d(\hat\zbf_i,\hat\zbf_j)
    \label{eq:cls}
\end{equation}
In both cases, $d(\cdot)$ is the Euclidean distance.

Finally, the clustering loss is computed as the ratio between the two distances, which when minimized encourages smaller intra-class and larger inter-class distances, given by
\begin{equation}
    \Lcal_\text{cls} = \frac{d_C+\epsilon}{d_R+\epsilon}, 
\end{equation}
where $\epsilon$ is a constant used for numerical stability. 
\begin{figure*}[t]
    \centering
    \includegraphics[width=\textwidth]{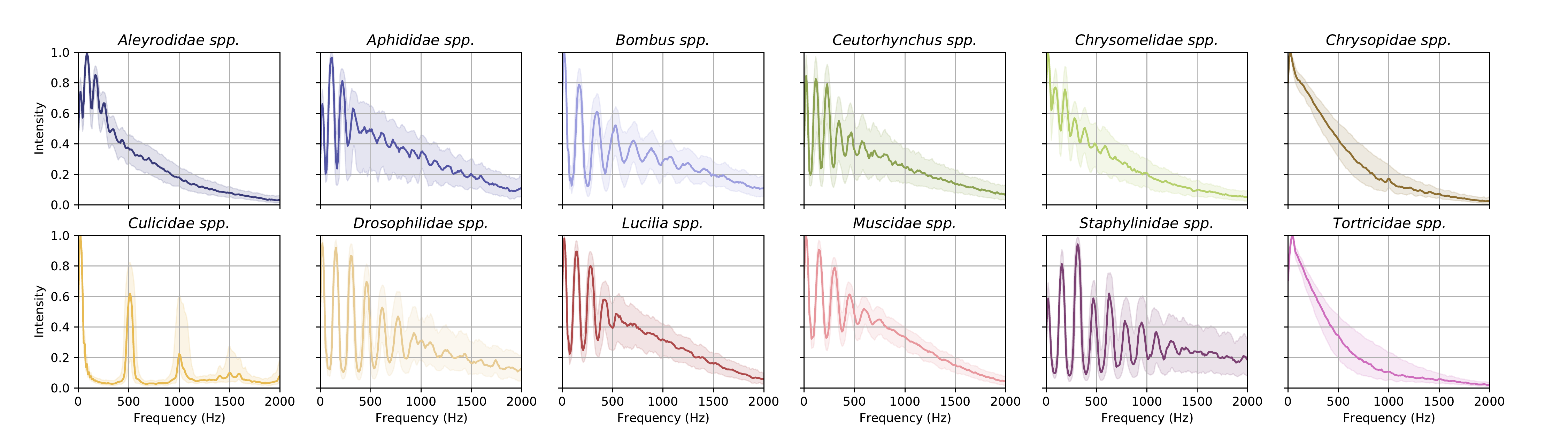}
    \caption{The median wing-beat frequency (WBF) spectra estimated from each labelled species in the input data, $\xbf_i \in \Rm^{F}$. The shaded areas indicate the inter-quartile range, between $25\%$ and $75\%$. The mosquitoes (\textit{Culicidae spp.}) have the highest WBF and the moths (\textit{Tortricidae spp.} the lowest. The weevils, (\textit{Ceutorhynchus spp.}) have a large variation around their fundamental WBF. All recordings are log-transformed and individually normalized.}
    \label{fig:avg_wbf}
    % \vspace{-0.25cm}
\end{figure*}

\subsection{Training of dynamic $\beta$-VAE}

The final objective of the dynamic $\beta$-VAE with semi-supervision in Eq.~\eqref{eq:final_obj} has three components which are introduced during the training in three successive stages:
\begin{enumerate}
    \item Warm-up phase ($\beta=0,\gamma=0$): In this phase, the model primarily learns to reconstruct the input data similar to bottleneck autoencoders. 
    \item Regularized phase ($\beta>0,\gamma=0$): In this phase, the dynamic control of $\beta$ sets in which smooths the learned latent space without deterioration in the quality of reconstructions.
    \item Semi-supervised phase ($\beta>0,\gamma>0$): In this phase, the learned latent space is steered to favour the downstream task of clustering. 
\end{enumerate}

\section{Experiments \& Results}
\begin{figure*}
\centering
  \includegraphics[width=\textwidth]{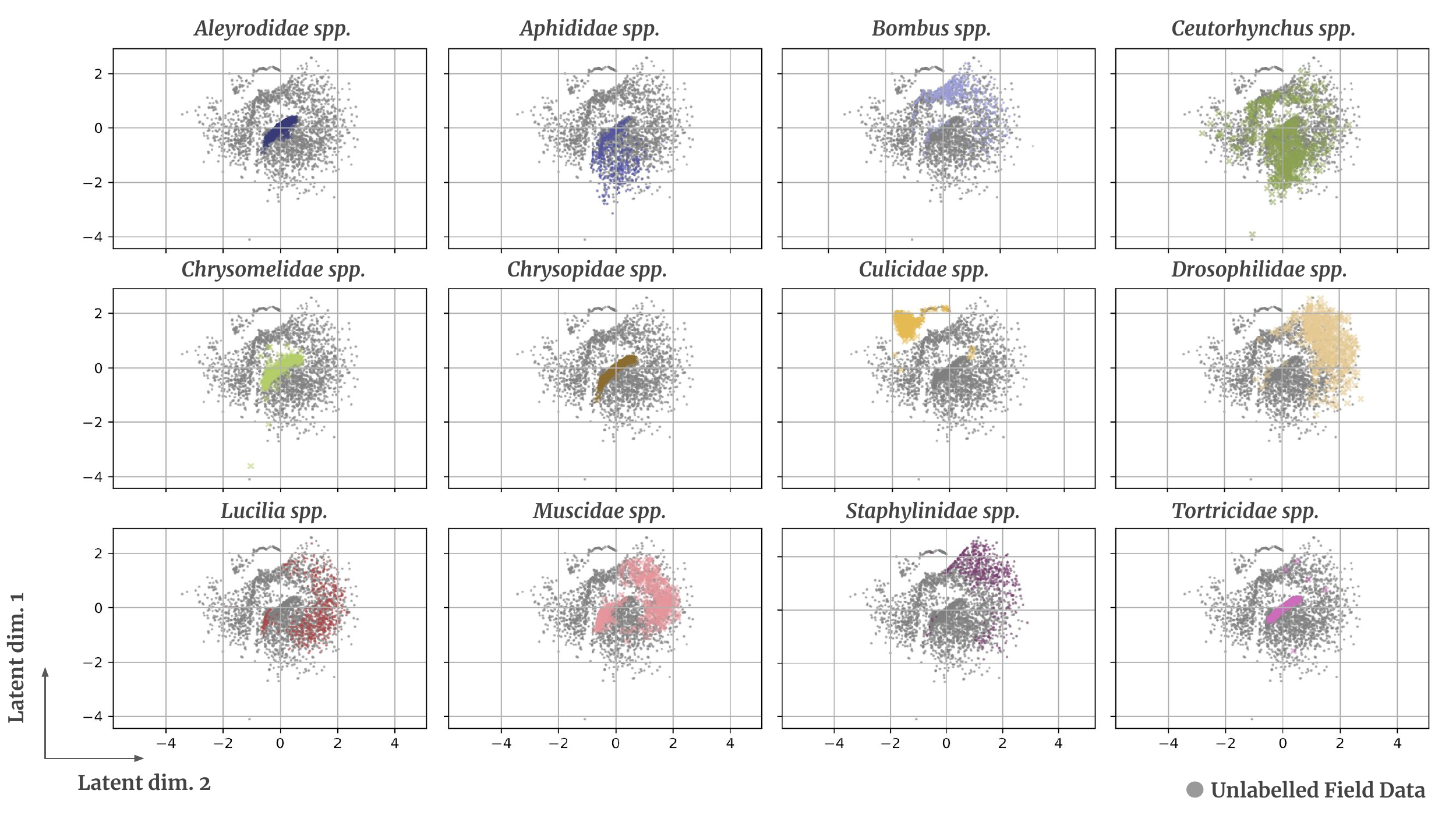}
  \caption{Latent representation of optically recorded insect wingbeat frequency spectra. The proposed dynamic $\beta$-VAE is able to cluster unlabelled field recordings into compact clusters. Evaluated on labelled data, most species form compact clusters, as shown with different colours for each of the 12 named species groups. Closely related species groups, such as the \textit{Lucila spp.} and \textit{Muscidae spp.} are partially overlapping but well separated from more distant species groups, such as the \textit{Bombus spp.}.}
  \label{fig:teaser}
\end{figure*}

The main objective of the proposed dynamic $\beta$-VAE is to cluster unlabelled insect spectra into plausible clusters that could correspond to unique species. To evaluate the performance of the model, we use real data collected from field instruments and compare the model's performance in different settings. Details of the data and experiments are presented in this section.

% \vspace{-0.25cm}
\subsection{Data Collection \& Pre-processing}

The insect data were recorded with a novel instrument from FaunaPhotonics, which uses a similar principle at close- and long ranges as the methods described in \cite{kirkeby11advances, gebru2018multiband, brydegaard2020lidar, brydegaard2014advantages}. In the current implementation, an air volume is illuminated with infrared light in two spectral bands at $808$ nm and $975$ nm. The back-scatter of any object passing through a $20$ l volume within $1$m of the sensor is recorded onto a photo diode quadrant detector. As insects fly past, the optical cross section varies with their WBF. This yields a modulated time series, sampled at $20$ kHz with a bandwidth from $0$ to $5$ kHz. As signals from any non-insect object passing through the volume are also recorded, a CNN trained with manual labels was used to filter out insect recordings from rain and dust etc. For simplicity, the multi spectral time series were reduced to one dimension by calculating the average Welch power spectra \cite{welch1967use} over both spectral bands in $F=193$ bins between 0 and 2kHz. Finally, the data was log-transformed and individually normalized by the maximum of each spectrum.

% Describe laboratory and field setups
The unlabelled data were recorded from March to November 2020 in various biotopes in the Öresund region in southern Scandinavia and $N=40,000$ insect recordings were randomly selected for the experiments with $F=193$ features after the WBF pre-processing. 

Additionally, data encompassing 12 different species groups were labelled one species at a time in closed cages in Copenhagen, Denmark. From this data, 
$6000$ insect recordings ($15\%$ of the unlabelled data) were randomly selected and added to the labelled training set. For each species group, this resulted in $500$ labelled recordings to be used in the semi-supervised mode. The average WBF spectra for each labelled species group is shown in Figure \ref{fig:avg_wbf}.

In order to validate the clustering ability of the different models, $8$ out of the $12$ labelled species were included in computing the clustering loss, $\mathcal{L}_{cls}$ in Eq.~\eqref{eq:cls}, in the semi-supervised setting. The remaining four labelled species were used as test set for validating the clustering accuracy.

% \vspace{-0.2cm}
\subsection{Experimental set-up}
The dynamic $\beta$-VAE was evaluated in unsupervised and semi-supervised modes to obtain latent representations, which were clustered using K-means {\cite{lloyd1982least}}. Their clustering performance was compared with {the baseline methods: PCA, Kernel-PCA, HCA using the standard implementations in sklearn \cite{pedregosa2011scikit} and a conventional VAE} on the same data. The encoder neural network $q_\theta(\zbf|\xbf)$ consists of 9 fully connected layers, with rectified linear unit (ReLU) activation (except for the last layer). The encoder predicts the mean and the variance of the approximate posterior distribution. The decoder neural network $p_\phi(\xbf|\zbf)$  is implemented with 10 fully connected layers and ReLU activation (except the last layer, which has sigmoid activation). The VAE uses a bottleneck $L=2$ to create the latent representation. {The model layout was developed on a independent unlabelled dataset recorded at a different location and was gradually expanded until reconstructions were sufficiently good. In order to visualize the latent representation, the size of the bottleneck of the model (latent dimension) was limited to two.} Details of the network architecture are reported in Table~\ref{tab:model}.

Both the unsupervised and semi-supervised models were run five times on random training and test splits. A random subset of $3000$ recordings from the dataset were used as the test set. 

After each training run the latent representation of the unlabelled test set was clustered using K-means method. The appropriate number of clusters $K$ were automatically selected by the maximum average silhouette score \cite{rousseeuw1987silhouettes} from a range of $5-50$. As we expect the unlabelled data to consist of at least $5$ distinct species, we incorporate this as prior information in choosing the range of clusters. For comparison, the data were also clustered into the same range of clusters using {PCA, Kernel-PCA (implemented with sigmoid kernels) and HCA (implemented with complete linkage).}

The final evaluation was done by comparing the automatically identified clusters with the four labelled test species. The automatically found clusters were compared with the labelled data using the adjusted rand index (ARI) \cite{rand1971objective} and adjusted mutual (AMI) information score \cite{vinh2010information}. These metrics are useful to compare clusters obtained in unsupervised settings, as they are agnostic to labels and only focus on the similarity between members within the clusters.

\begin{table}[h]
\footnotesize
\centering
    \caption{Network architecture of the implemented dynamic $\beta$-VAE showing the number of hidden units per layer (H) and the non-linear activation functions per layer in the encoder and decoder parts of the network. (RL: Rectified Linear Unit. SG: Sigmoid.)}
    \label{tab:model}
    \begin{tabular}{lccccccccccc}
    \toprule
         \textbf{\#} &  & \textbf{1} & \textbf{2} & \textbf{3} & \textbf{4} & \textbf{5} & \textbf{6} & \textbf{7} & \textbf{8} & \textbf{9} & \textbf{10} \\
        \midrule
        \textbf{Enc.} & H & 193 & 128 & 128 & 64 & 32 & 16 & 8 & 4 & 2+2 & --\\
        & Act. & RL & RL & RL & RL & RL & RL & RL & RL & --  & --\\
        \midrule
        \textbf{Dec.} & H & 2 & 4 & 8 & 16 & 32 & 32 & 64 & 128 & 128 & 193 \\
         & Act. & RL & RL & RL & RL & RL & RL & RL & RL &  RL & SG \\
         \bottomrule
    \end{tabular}
    \vspace{-0.5cm}
    \end{table}
    
\begin{figure*}[t]
\centering
     \begin{subfigure}{0.3\textwidth}
         \includegraphics[width=\textwidth]{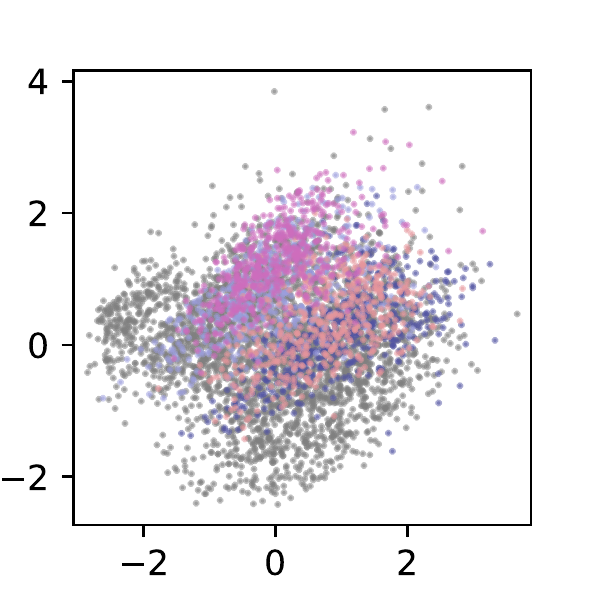}
         \caption{PCA}
         \label{fig:lat_pca}
     \end{subfigure}
    % \hfill
     \begin{subfigure}{0.3\textwidth}
         \includegraphics[width=\textwidth]{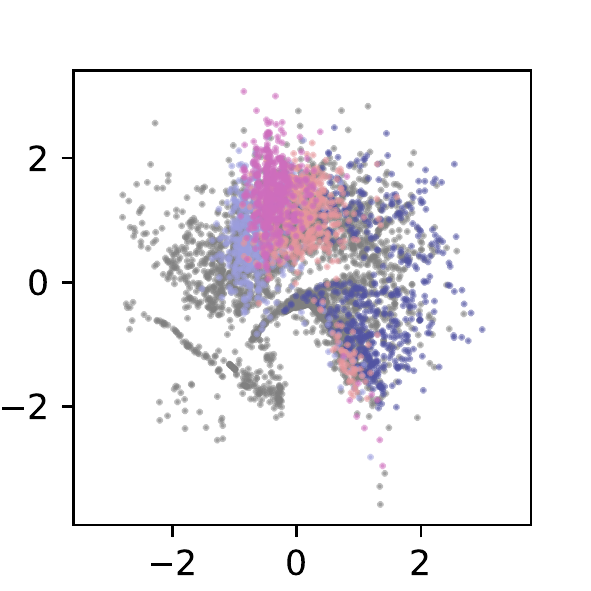}
         \caption{Kernel-PCA}
         \label{fig:lat_kpca}
     \end{subfigure}
    % \hfill
     \begin{subfigure}{0.3\textwidth}
         \includegraphics[width=\textwidth]{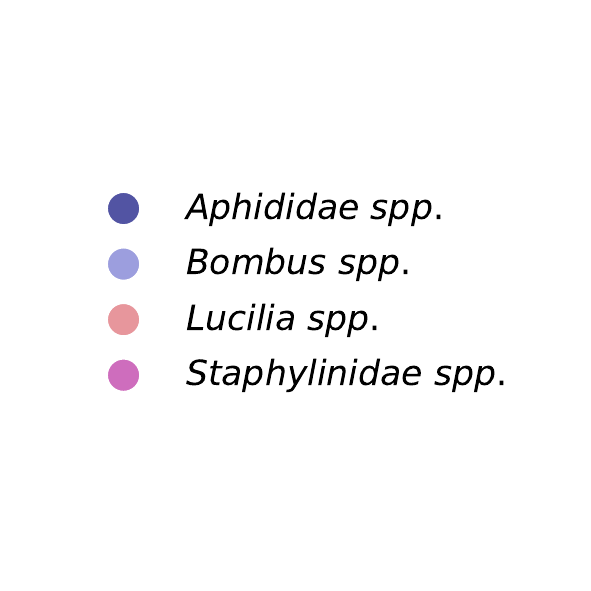}
         \label{fig:com_legend}
     \end{subfigure}
    % \hfill
     \begin{subfigure}{0.3\textwidth}
         \includegraphics[width=\textwidth]{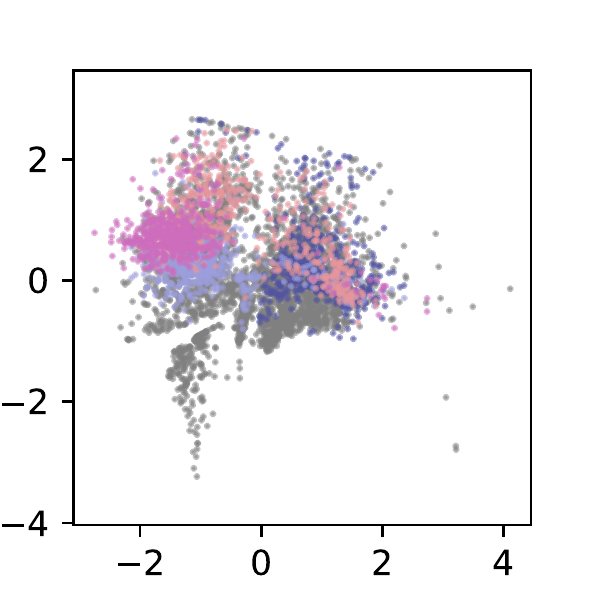}
         \caption{VAE}
         \label{fig:lat_vae}
     \end{subfigure}
    % \hfill
     \begin{subfigure}{0.3\textwidth}
         \includegraphics[width=\textwidth]{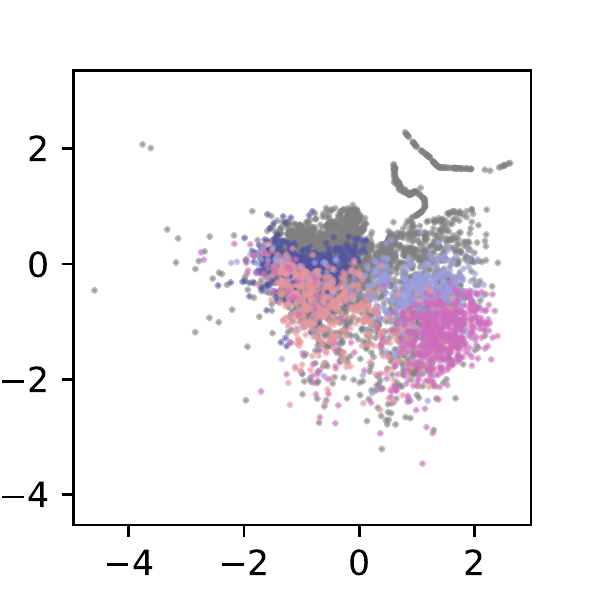}
         \caption{Unsupervised $\beta$-VAE}
         \label{fig:lat_usvava}
     \end{subfigure}
     \begin{subfigure}{0.3\textwidth}
         \includegraphics[width=\textwidth]{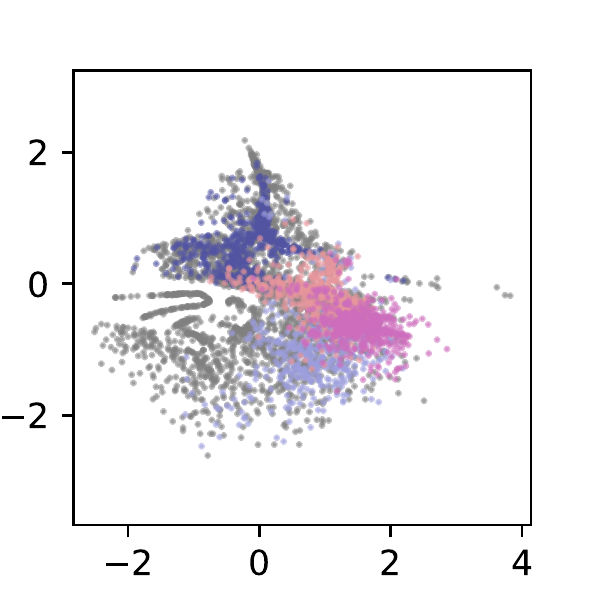}
         \caption{Semi-supervised $\beta$-VAE}
         \label{fig:lat_ssvae}
     \end{subfigure}
    \hfill
     \caption{Example latent representation of unlabelled field data and the four test species. While all methods map the high dimensional input data to a two dimensional feature space, the {dynamic $\beta$}-VAEs creates clusters with less overlap between the species groups. The inclusion of $10\%$ labelled data for training further improves the results, yielding denser clusters with less overlap than the unsupervised $\beta$-VAE.}
     \label{fig:model_comp}
\end{figure*}

\subsection{Model parameters \& hyperparameters}
The dynamic $\beta$-VAE has several tunable model parameters, as seen in Eq.~\eqref{eq:beta-vae} to Eq.~\eqref{eq:w3}. These model parameters were tuned on an independent dataset, collected with the identical instrumentation but at a different location. The data had a similar distribution as the data used in this work and we obtained: $a=0.2$ and $b=0.05$, $w_1=w_2=1.2$, $w_3=0.9$ and $w_4=1.1$. These parameters were found to be sufficiently robust on the dataset used in this work without any fine-tuning. 

All models were implemented in PyTorch \cite{paszke2019pytorch} and trained for $5000$ epochs using the Adam optimizer \cite{kingma2015adam} with a learning rate of $10^{-3}$. The models were trained on Nvidia GTX 1050 graphics processing unit with $4$ GB memory with a batch size of $256$. A decision to adapt $\beta$ was taken every fifth epoch to avoid random fluctuations. The scaling of the clustering loss, $\gamma$, in the semi-supervised mode was cycled between $0.01$ and $0.2$ every $100$ epochs.

%% This text should be blue!!!
\begin{table}[h]
\small
\centering
    \caption{Aggregated results from 5 repetitions of each method. The unsupervised model performs better than the classical models and adding the labelled data further improves clustering of unlabelled data. The median number of automatically determined clusters (K) are also reported. (ARI and AMI scores: higher is better.}
    \label{tab:results}
    \begin{tabular}{lccc}
    \toprule
         {\bf Models} & {\bf $K$} & {\bf ARI-score} & {\bf AMI-score} \\
         \midrule
         PCA            & $5$ & $0.15 \pm 0.02 $ & $0.21 \pm 0.01 $ \\
         K-PCA          & $5$ & $0.17 \pm 0.02$ & $0.22 \pm 0.01$ \\
         HCA            & $16$ & $0.11 \pm 0.06$ & $0.21 \pm 0.10$  \\
         VAE            & $5$ & $0.14 \pm 0.09$ & $0.20 \pm 0.09$ \\
         $\beta$-VAE    & $7$ & $0.25 \pm 0.02$ & $0.34 \pm 0.03$ \\
         $\beta$-VAE (semi-sup.)    & $6$ & $0.28 \pm 0.05$ & $0.37 \pm 0.05$ \\
         \bottomrule
    \end{tabular}
    \vspace{-0.5cm}
    \end{table}

% Cartwheel
\begin{figure*}[t]
    \centering
    \includegraphics[width=0.9\textwidth]{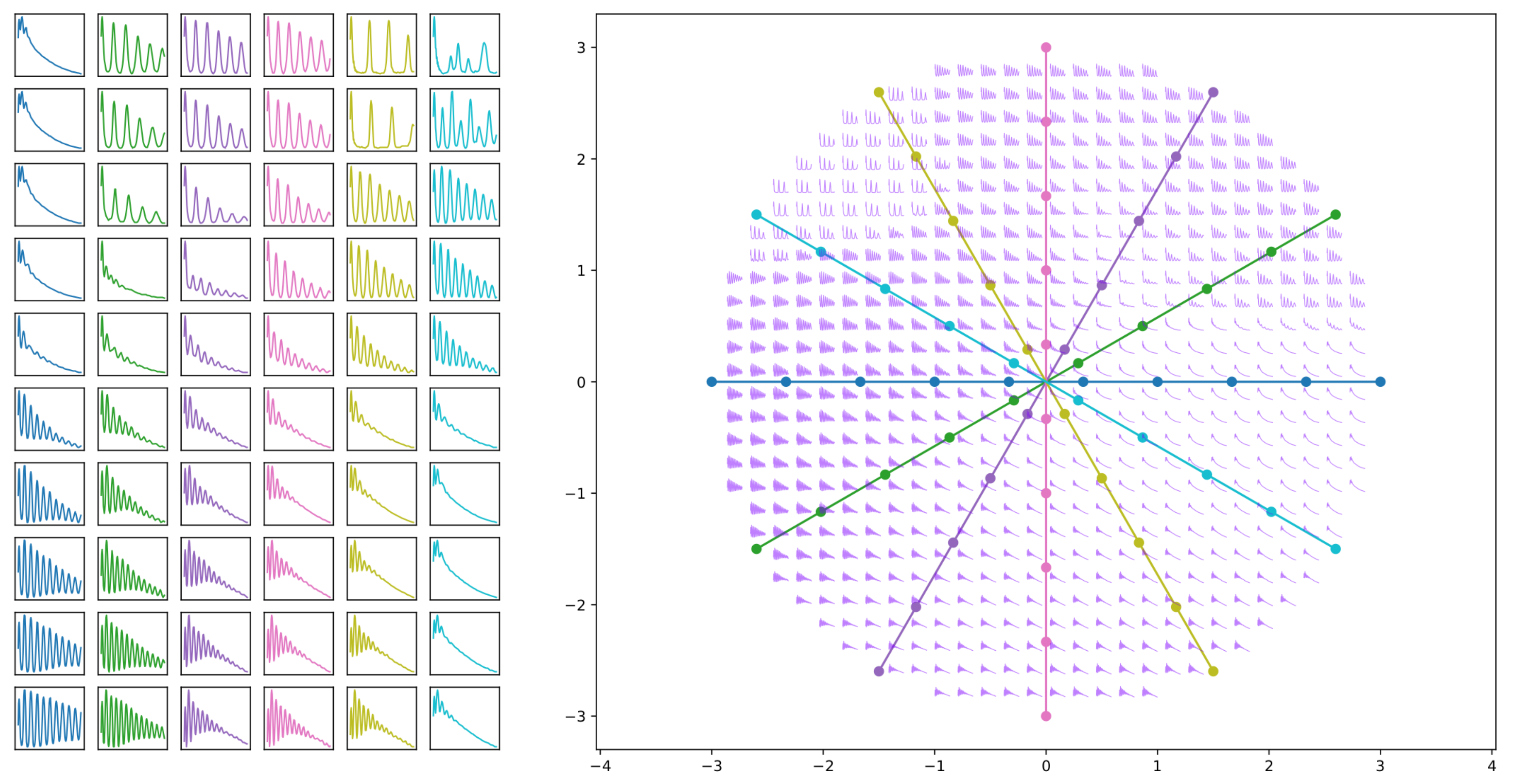}
    \caption{Latent space cart-wheel visualization. Samples from the latent representation of the semi-supervised dynamic $\beta$-VAE are shown with decoded latent samples when the lines are traversed. While the two dimensions do not appear fully disentangled, the latent space is regularized and transitions between various areas are smooth and gradual.}
    \label{fig:latent_wheel}
    
\end{figure*}
\begin{figure*}[t]
     \centering
     \begin{subfigure}[t]{0.49\textwidth}
         \centering
         \includegraphics[width=\textwidth]{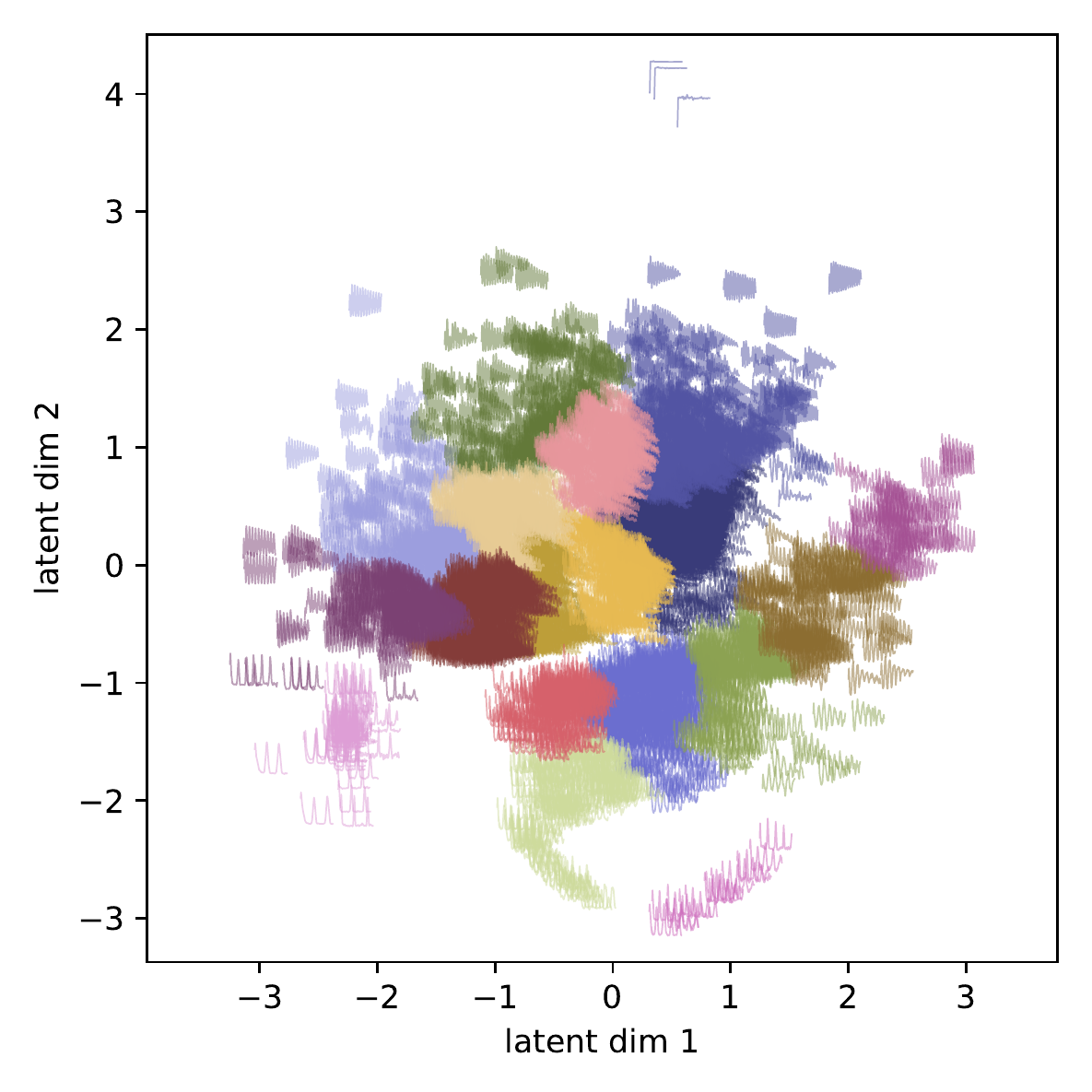}
         \caption{Latent representation}
         \label{fig:15_latent}
     \end{subfigure}
    \hfill
     \begin{subfigure}[t]{0.49\textwidth}
         \centering
         \includegraphics[width=\textwidth]{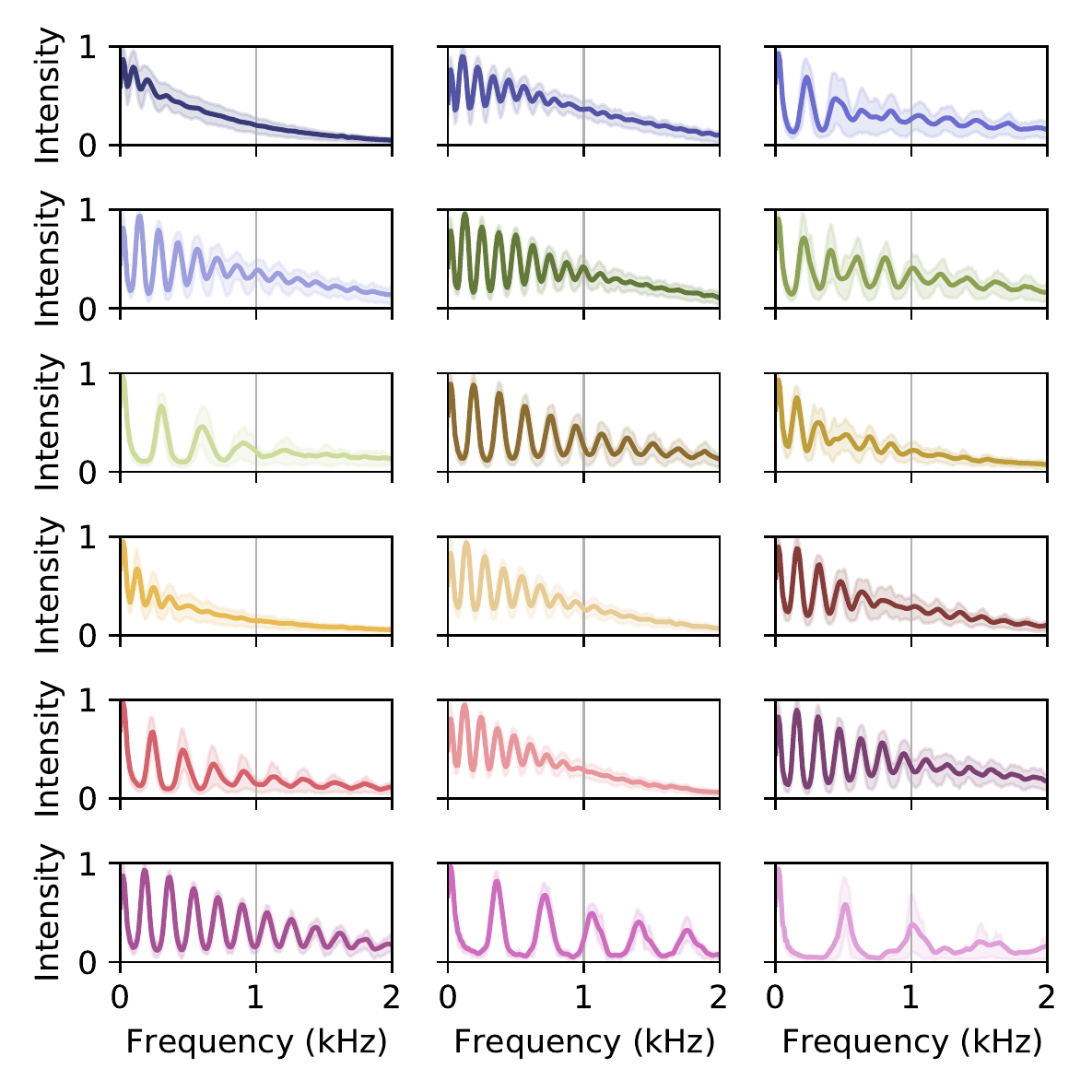}
         \caption{Average WBF spectra and IQR in each cluster}
         \label{fig:15_spectra}
     \end{subfigure}
     \caption{K-means clustering in the latent representation of unlabelled field data from the  semi-supervised $\beta$-VAE into 15 clusters. The model is capable of generating a low dimensional space where similar insect recordings are represented clustered together. The lower right cluster in \textbf{b)} are likely to contain mosquitoes due to their high WBF.}
     \label{fig:Unsupervised_clusters}
\end{figure*}

\subsection{Results}
The clustering performance on the labelled test set for the unsupervised and semi-supervised instances of the dynamic $\beta$-VAE is presented in Table~\ref{tab:results}.

{The dynamic $\beta$-VAE performs} better than {the baselines} in the ARI- and AMI-scores which quantifies the intra-class cohesion and inter-class separability. {While HCA have been successfully used to identify groups of similar insects by other groups previously~\cite{brydegaard2020lidar}, it has the lowest ARI score.} With semi-supervision the dynamic $\beta$-VAE further improves upon its unsupervised clustering scores, and the improvements compared to the conventional models are more pronounced. {The nonlinear kernel-PCA does not show any drastic improvements over conventional PCA.} The low dimensional representations of the test species for each model {are} shown in Figure \ref{fig:model_comp}. While the different species form largely separable and homogeneous clusters in all methods, they are relatively more compact in the semi-supervised implementation.

In the results presented in Table \ref{tab:results}, the appropriate number of clusters found in the unlabelled test set data is also reported. The number of clusters, $K$, was automatically chosen to maximize the average silhouette score \cite{rousseeuw1987silhouettes}. 

An example of the latent representation from all 12 labelled species groups by the unsupervised instance is shown in Figure \ref{fig:teaser}. All species generate dense, but partly overlapping, clusters except the weevils (\textit{Ceutorhynchus spp.}), and to some degree the fruitflies (\textit{Drosophilidae spp.}) which form sparser clusters. 

The latent space obtained by the semi-supervised $\beta$-VAE on the unlabelled test set is shown in Figure \ref{fig:15_latent}. Using  $K=15$ the data is color coded by cluster and the average spectra from each cluster is shown in Figure \ref{fig:15_spectra}. 

\section{Discussions}
In this work, we introduced a dynamic $\beta$-VAE in order to achieve a good trade-off between the reconstruction- and regularization loss terms {by performing online adjustment of the $\beta$ term}. The proposed $\beta$ dynamics result in useful latent representations for the downstream clustering task. Our experiments demonstrate the ability of the model to map high-dimensional insect data into a well regularized latent representation where phylogentic groups are distinguishable.

\subsection{Generalization of the $\beta$ dynamics}
{
The primary objective of using the $\beta$ dynamics in Eq.~\eqref{eq:dyn-vae} is to perform {\em online} adjustment of the scales of reconstruction and regularization terms based on their instantaneous values while taking the previous optima into account. The specific formulation of these control mechanisms in Eq.~\eqref{eq:w1}-~\eqref{eq:w3} force the model optimization to not deviate from the previous optimal solutions. The terms comprising $min$ over $(:t-1)$ epochs in Eq.~\eqref{eq:w1} and~\eqref{eq:w2} provide a form of memory of the previous local optima. The trade off between long and short term memory of the losses and the corresponding optima help the model to steer towards more global optima. These equations provide a sufficiently general formulation for adjusting $\beta$ as they are only dependent on the two loss components. Further, one can also envision a learnable neural network with long short-term memory (LSTM) that can perform this dynamic control in a recurrent neural network type formulation of a closed loop control system~\cite{hochreiter1997long}.}

\subsection{Influence of the $\beta$ {and $\gamma$} parameters}

As seen in Figure \ref{fig:progression}, the initial effect of a dynamic $\beta$ is similar to $\beta$-annealing, where $\beta$ gradually increases during training in order to prevent posterior collapse \cite{bowman2015generating}. However, the key difference with $\beta$ annealing is that the rate of annealing is not predetermined, as the $\beta$ dynamics described in Sec.\ref{sec:dyn} enables self-regulation of $\beta$. This is witnessed during the latter part of training, where $\beta$ repeatedly adapted either by increasing or decreasing its value dependent on the changes in the reconstruction- and regularization terms. This behaviour is similar to what is reported with the cyclic $\beta$-VAE \cite{fu2019cyclical} where the {\em shakeup} often allows the model to obtain new global minimum loss. The similarity to the cyclic $\beta$-VAE is further enhanced by automatically increasing $\beta$ when there has been no change for a large number of epochs (500 in our case). However, unlike a cyclic $\beta$-VAE, $\beta$ is only cycled if it has reached a stationary condition and $\Delta_{reg}$ and $\Delta_{rec}$ are within limits. This helps both the unsupervised- and semi-supervised instances of dynamic $\beta$-VAE to latch on the historical minimum of both $\mathcal{L}_{rec}$ and $\mathcal{L}_{reg}$. 

The $\beta$ dynamics introduced in Sec. \ref{sec:dyn} is also similar to adaptive strategies used in models such as the controlVAE \cite{shao2020controlvae}. A controlVAE stabilizes the model by adjusting $\beta$ to keep the regularization loss (KLD) at a constant level. However, finding an appropriate KLD level can be difficult as it could vary across datasets and downstream tasks. In contrast, the dynamic $\beta$-VAE keeps the model stable by constantly comparing both $\mathcal{L}_{rec}$ and $\mathcal{L}_{reg}$ with their historical minima. A gain on either loss term, at the expense of the other, is counteracted by adjusting $\beta$. This self-regulating $\beta$ dynamics that is not dependent on fixing KLD value is an advantage with our formulation.

{Including the additional loss term scaling term $\gamma^{(t)}  \mathcal{L}_{\text{cls}}$ in Eq.~\eqref{eq:final_obj} further improved the clustering performance of the model. In this implementation $\gamma$ was cycled between 0.01 and 2 in order to keep the contribution from $\mathcal{L}_{\text{cls}}$ in a similar range as $\mathcal{L}_{\text{rec}}$ and $\mathcal{L}_{\text{reg}}$. A logical next step could be to expand Eq.~\eqref{eq:dyn-vae} to include a dynamically adjusted $\gamma^{(t)}  \mathcal{L}_{\text{cls}}$ term. This would however make the model less generalized to other tasks.}

\subsection{Performance comparison}
Comparing the performance between the models reported in Table~\ref{tab:results}, the {dynamic $\beta$-VAE} perform better than the baseline models. Adding $15$\% of labelled data from 8 different species to the training set further improves the clustering performance. {While using HCA on high dimensional frequency spectra have been successfully used to identify mosquito clusters in field data \cite{brydegaard2020lidar} and biodiversity evaluation \cite{kouakou2020entomological}, our results show better performance for PCA + Kmeans. While the Kernel-PCA generally produced more heterogeneous latent distributions, it did not show any significant improvements over the standard PCA.}

{The non adaptive $\beta$-VAE showed large variation in its performance but was on average comparable with the conventional methods. The dynamic $\beta$-VAEs kept $\beta < 1$ during most of the training, as exemplified in Figure \ref{fig:progression}, and since a higher $\beta$ term favours a well generalised latent space over good reconstructions, a reduction in clustering performance could be expected.
Additionally, the non adaptive VAE was more cumbersome to train as the model collapsed frequently during training.}

\subsection{Selection of number of clusters}
In this work, the appropriate number of clusters were automatically selected by maximizing the average silhouette score. However, when manually evaluating the average silhouette score and comparing it with commonly used empirical measurements, such as the elbow method \cite{thorndike1953belongs} and intra-cluster sum-of-squares, a user might typically identify a higher number of clusters. 
Having more clusters yield more similar recordings within each cluster. An example is show in Figures \ref{fig:15_latent} and \ref{fig:15_spectra} where the number of clusters were manually selected. As in previous work by lidar-entomologists \cite{brydegaard2020lidar}, some species groups can be identified from these clusters at this level by comparing the average spectra of each cluster with known data. For example a cluster of possible mosquitoes can be identified by their high wing-beat frequency the lower right corner of Figure \ref{fig:15_spectra}.

With the $3000$ randomly chosen insect recordings from multiple sites during summer, we expect the total number of species represented in the test set to be one or several orders of magnitude larger. We tested a range of $5-50$ clusters as even reasonably coarse clustering will be useful for quantifying biodiversity. Once fully deployed on a network of insect sensors, a dataset recorded in an environment with high biodiversity could yield more clusters than a dataset captured in a biologically poor environment. This would allow the automated and optically recorded insect data to be correlated with conventional monitoring methods and greatly improve the ability to monitor insect biodiversity at scale.

\subsection{Computation time and inference}
Computation time for PCA + K-Means is significantly shorter compared to HCA on the full spectra \cite{brydegaard2020lidar, kouakou2020entomological}. While the initial training of the dynamic $\beta$-VAE takes a few hours depending on the number of epochs, once trained, the inference time for the model is comparable with that of using PCA + K-Means. Once deployed in the field, the dynamic $\beta$-VAE model is not expected to be retrained regularly but to be used as a dimensionality reduction method. Therefore, inference time is a more important metric than the initial computation time. 

\subsection{Exploring the latent space}
Samples generated from the latent space of the semi-supervised model are shown as a {\em latent space cart-wheel} in Figure \ref{fig:latent_wheel}. Traversing different lines in the latent space results in samples that smoothly transition between different spectra types. As a side note, we point that the two latent features do not appear to be entirely disentangled; this is manifested as dense islands and sparse spaces of spectra in the latent space. For our downstream clustering task, fully disentangled features are not required. However, one could introduce an additional loss component that enforces orthogonality between the different latent dimensions to achieve improved disentanglement. 

The latent representation of the unsupervised model can be further validated by comparing Figure \ref{fig:teaser} with the average spectra of each group, shown in Figure \ref{fig:avg_wbf}. Species groups with similar spectra, such as \textit{Aleyrodidae spp.}, \textit{Aphididae spp.} \textit{Tortricidae spp.} and \textit{Chrysopidae spp.} are positioned in similar areas. Similarily, all dipterans (\textit{Lucilia spp.}, \textit{Muscidae spp.} and \textit{Drosophila spp.}) have overlapping clusters except the mosquitoes (\textit{Culicidae spp.}) which have a much higher WBF and are more isolated. In Figure \ref{fig:teaser}, the model performs less well on the weevils (\textit{Ceutorhynchus spp.}) compared to the other species. It is likely due to a larger variation in their WBF than for the other groups. 

\section{Conclusions}
In this work, we have presented, to our knowledge, the first VAE designed for clustering of optically recorded insect signals. The dynamic $\beta$-VAE was developed in order to achieve a stable model, optimizing both reconstruction- and regularization terms. When trained on unlabelled data recorded during field conditions, the model is able to automatically create meaningful clusters. The unsupervised clustering performance was validated with labelled data collected in controlled conditions showing promising results.
By using $15$\% of labelled data during the training process for semi-supervision, this clustering performance is further improved. Already at this stage, it is possible to extract easily identifiable insect groups such as mosquitoes from the automatically identified clusters and we expect this capability to grow as the collection of labelled reference data continues.

The future aim of our work is to further improve automatic identification of the number of clusters. In the {near future}, the model will be deployed on several sensors in the field, and estimated cluster sizes and distributions will be compared to conventional methods. This will greatly improve monitoring possibilities and decision support tools for entomologists and agronomists.
In order to mitigate the trend of declining insect communities, the first step is to ensure adequate data collection. In the near future, we believe methods based on the proposed dynamic $\beta$-VAE will be useful to quantify and, thus, help conserve insect biodiversity.
\section*{Acknowledgments}
Ankit Kariyaa, Inger Kappel Schmidt and Laurence Still for proof reading.
Marta Montaro and teammates for labelled data recordings.
Erik B Dam and Christoffer Grønne for helpful discussions. 

\section*{Competing interests}
Klas Rydhmer is employed at FaunaPhotonics. We declare that this have not affected the reported results or interpretations in any way. Raghavendra Selvan declares no competing interests.

\section*{Funding}
This work was supported by the Innovation Fund Denmark.
% \appendix

 \bibliographystyle{elsarticle-num} 
 \bibliography{references}





% \end{thebibliography}
\end{document}